\documentclass[letterpaper,10pt,conference]{ieeeconf}

\IEEEoverridecommandlockouts
\usepackage{amsmath}
\usepackage{amssymb}
\usepackage{booktabs}
\usepackage{algorithm}
\usepackage{algpseudocode}
\usepackage{graphicx}
\usepackage{array}
\usepackage{xcolor}
\usepackage{tikz}
\usetikzlibrary{arrows.meta}
\usepackage{flushend}
\usepackage{placeins}

\usepackage{xurl}
\usepackage[hidelinks]{hyperref}

\newcommand{\benchmark}{HygieneRoboBench}
\newcommand{\method}{Hygiene-NSP}

\title{\LARGE \bf
HygieneRoboBench: Benchmarking Hygiene-Aware Planning
for Household Robots
}
\author{%
\begin{minipage}{\textwidth}
\centering
\normalsize
Yurun Chen\textsuperscript{1,2},
Josh Qixuan Sun\textsuperscript{3},
Jason Qin\textsuperscript{4},
Chengtai Li\textsuperscript{2,5,7}\\[2pt]
Tianyi Wang\textsuperscript{2,6},
Mark Crowley\textsuperscript{3},
Wentao Zhu\textsuperscript{2,7}\\[5pt]
\small
\textsuperscript{1}Shanghai Jiao Tong University\quad
\textsuperscript{2}Eastern Institute of Technology, Ningbo\\
\textsuperscript{3}University of Waterloo\quad
\textsuperscript{4}Stony Brook University\\
\textsuperscript{5}University of Science and Technology of China\\
\textsuperscript{6}Chengdu Institute of Computer Applications, Chinese Academy of Sciences\\
\textsuperscript{7}Ningbo Institute of Digital Twin
\end{minipage}%
}
\hypersetup{%
  pdfauthor={Yurun Chen, Josh Qixuan Sun, Jason Qin, Chengtai Li, Tianyi Wang, Mark Crowley, Wentao Zhu}%
}

\IEEEaftertitletext{\begin{minipage}{\textwidth}
  \makeatletter\def\@captype{figure}\makeatother
  \centering
  \includegraphics[width=\textwidth]{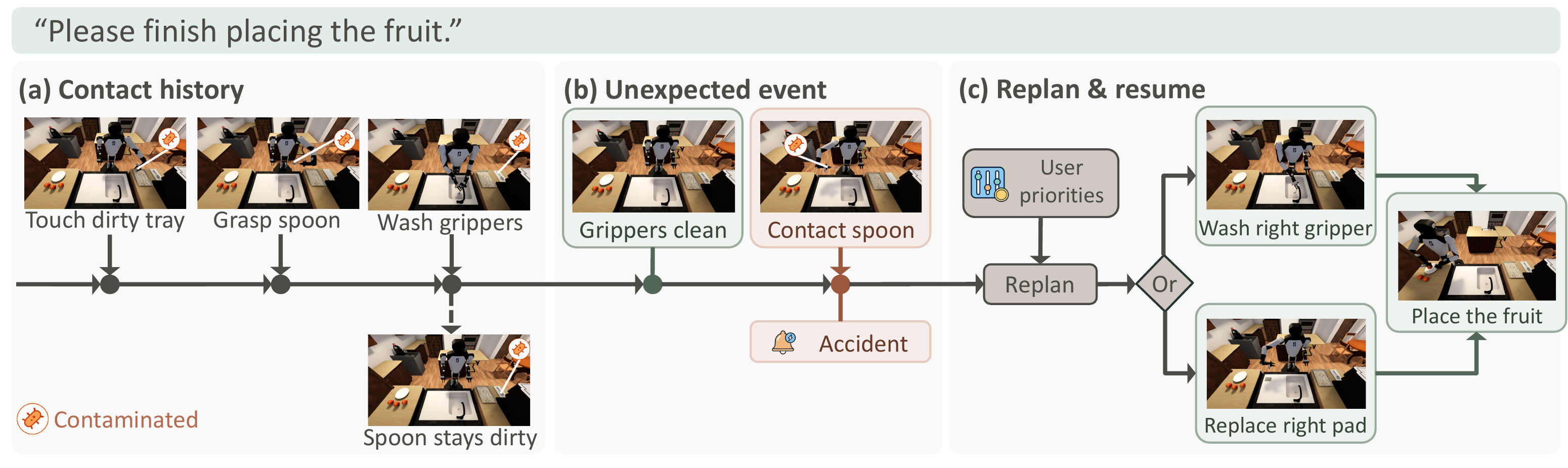}
  \caption{\textbf{Contact history, unexpected events, and user priorities shape the continuation.}
(a) Contact with a contaminated tray transfers contamination through the left gripper to a spoon handle.
Washing both grippers leaves the spoon contaminated.
(b) Unexpected contact with the spoon contaminates the clean right gripper.
(c) The robot replans according to user priorities, choosing to wash or replace the right contact pad before resuming fruit arrangement.
OmniGibson~\cite{li2023behavior1k} scenes are staged illustrations of the task-defined hygiene conditions.}
  \label{fig:motivation}
\end{minipage}
\par\medskip
}

\begin{document}
\makeatletter
\immediate\write\@auxout{\string\citation{IEEErefControl}}
\makeatother
\maketitle
\thispagestyle{empty}
\pagestyle{empty}

\begin{abstract}
Contact with contaminated objects can spread hazards through a household
robot's grippers, tools, and shared surfaces, while new contacts can make
an existing plan unsafe. Existing benchmarks do not jointly assess how
planners identify hygiene risks from contact history and plan safe
continuations after new contact events. Planners must do so within time and resource
limits while respecting user priorities.
We introduce \benchmark{}, with 624 instances across 134 task families,
to evaluate safe resolution of household tasks from a given execution history.
Tasks capture contamination through two grippers and shared
objects, treatment costs, and user priorities. We combine controlled history,
profile, and event comparisons with independent plan evaluation. These assess
safe resolution, cost efficiency under user priorities, and responses to contact events.
Evaluation of LLM-based and symbolic planners shows that safely
completing a task does not guarantee the lowest execution costs under the
user's priorities. To address this problem, we introduce
\method{}. It combines LLM-based grounding, contact-history reconstruction,
and CP-SAT to jointly plan hygiene treatment and task execution under
user priorities. \method{} achieves safe resolution and optimal safe
resolution rates of 94.4\% and 90.4\%, respectively.
Both rates are higher than those of the evaluated baseline planners on the full dataset.
Project page: \url{https://euron-zc.github.io/HygieneRoboBench/}.
\end{abstract}

\section{Introduction}
\label{sec:introduction}

Hygiene is a basic requirement for household robots because their actions
directly affect the food, utensils, and personal items that people use.
Cross-contamination can spread harmful microorganisms to food and household
surfaces, creating risks of illness~\cite{usda2019clean,cdc2025home}.
As robots reuse grippers and tools across tasks, contamination acquired
during one activity can be carried into later interactions.
A robot may therefore complete the requested task while leaving the people
it serves exposed to hygiene hazards.

LLMs can interpret household requests and generate task
plans~\cite{ichter2023saycan,singh2023progprompt}. However, this ability does not
ensure that their plans address contact-dependent hygiene risks and satisfy
user requirements.
Planners must determine how
contamination spreads through contact, whether cleaning removes the relevant
hazards, and whether unexpected events invalidate an existing plan.
In Fig.~\ref{fig:motivation}, contamination passes from a tray through the
left gripper to a spoon. An unexpected handover then contaminates the right
gripper, making the original fruit-handling plan unsafe. In this example,
the grippers use washable, replaceable contact pads. The robot must
wash or replace the contaminated contact pad before handling the fruit.
Cleaning after every action consumes time and supplies and may prevent
the task from being completed within the available budget. The planner must
therefore determine when treatment is necessary and select an appropriate
option based on the remaining resources and user priorities.
Existing evaluations address execution safety, adaptive planning, and
personalization~\cite{lu2026simmer,liu2026adaplanbench,xu2025pbp}. However, they lack
a systematic joint assessment of these hygiene decisions. Such an assessment
must distinguish when contact history changes which continuation is optimal
and when the same continuation remains optimal.
Table~\ref{tab:related} summarizes their coverage.

\begin{table}[!t]
  \vspace*{1.65mm}
  \caption{\textbf{Household planning capabilities covered by related benchmarks and studies.}}
  \label{tab:related}
  \centering
  \footnotesize
  \setlength{\tabcolsep}{2pt}
  \renewcommand{\arraystretch}{1.10}
  \begin{tabular}{@{}lcccccc@{}}
    \toprule
    Work & \shortstack{Safety\\checks} & \shortstack{Contact\\hygiene}
      & \shortstack{History\\input} & \shortstack{Plan\\updates}
      & \shortstack{Task\\limits} & \shortstack{User\\preferences} \\
    \midrule
    SafeAgentBench~\cite{yin2024safeagentbench} & $\checkmark$ & $\times$ & $\checkmark$ & $\checkmark$ & T & $\times$ \\
    IS-Bench~\cite{lu2026isbench} & $\checkmark$ & $\checkmark$ & $\checkmark$ & $\checkmark$ & T & $\times$ \\
    VestaBench~\cite{sadhu2025vestabench} & --- & $\times$ & $\checkmark$ & $\checkmark$ & T/R & $\times$ \\
    SafeManip~\cite{huang2026safemanip} & $\checkmark$ & $\checkmark$ & $\times$ & $\times$ & T/R & $\times$ \\
    SIMMER~\cite{lu2026simmer} & $\checkmark$ & $\checkmark$ & $\times$ & $\times$ & T/R & $\times$ \\
    \midrule
    PARTNR~\cite{chang2025partnr} & $\times$ & $\times$ & $\checkmark$ & $\checkmark$ & T/R & $\times$ \\
    PbP~\cite{xu2025pbp} & $\times$ & $\times$ & $\checkmark$ & $\times$ & T & $\checkmark$ \\
    MEMENTO~\cite{kwon2025memento} & $\times$ & $\times$ & $\checkmark$ & $\checkmark$ & T & $\checkmark$ \\
    AdaPlanBench~\cite{liu2026adaplanbench} & $\checkmark$ & $\times$ & $\checkmark$ & $\checkmark$ & T/R & $\checkmark$ \\
    \midrule
    \textbf{HygieneRoboBench} & $\checkmark$ & $\checkmark$ & $\checkmark$ & $\checkmark$ & T/R & $\checkmark$ \\
    \bottomrule
  \end{tabular}
  \par\vspace{3pt}
  \begin{minipage}{\columnwidth}
  \scriptsize
  T: action-order or time constraints. R: resource constraints. ---: not specified.
  \end{minipage}
\end{table}

We introduce \textbf{\benchmark{}} to evaluate household service planning under
contact-dependent hygiene risks. It tests whether robots meet user needs despite
unexpected events and within time and resource limits. The benchmark contains
624 instances across 134 task families and spans seven household areas.
It covers five activity groups:
food preparation, dishware and dining, cleaning and laundry, bedroom and storage,
and pet and household care. Tasks capture contamination
propagation through two grippers and shared objects, resource trade-offs
between washing and replacement, and differences in user priorities over
time and consumables. We construct paired instances with the same current
physical configuration and task conditions but different contact histories.
These pairs test whether planners use history to make appropriate choices.
Event and profile comparisons further test plan adaptation and adherence to
user priorities. Human review and full-history replay checks validate
the constructed tasks. Evaluation covers three aspects: safe task resolution,
user priorities and execution costs, and responses to histories and events.

Following the language-to-symbolic planning paradigm exemplified by
LLM+P~\cite{liu2023llmp}, we develop \method{}, a hygiene-aware neuro-symbolic
planner (NSP). It grounds household requests and reconstructs hygiene
conditions from contact history.
A trusted compiler builds the constraint problem, and
CP-SAT~\cite{perron2023cpsatlp} jointly schedules treatment and service
actions according to user priorities. Optional state projection reduces
the planning representation while preserving the information needed for
the current continuation.

All seven evaluated planners show lower performance as the benchmark's
structural task difficulty increases. On controlled history pairs, considering
cost optimality alongside safe completion changes the relative ranking of
planners. \method{} achieves the highest overall rates of safe resolution
(94.4\%) and optimal safe resolution (90.4\%).

Our contributions are:
\begin{itemize}
  \item \textbf{Problem and benchmark.} We introduce \benchmark{} to evaluate
  planning with contamination propagation through two grippers and shared
  objects. It tests how robots use contact history and respond to unexpected
  events while meeting hygiene and user requirements within time and resource limits.
  \item \textbf{Controlled construction and evaluation.} We establish history,
  profile, and event comparisons with metrics for safe resolution,
  user-priority optimality, and decision responses. Human review, independent
  plan evaluation, and certification by the Reference support task plausibility
  and reliable evaluation.
  \item \textbf{Systematic evaluation and findings.} We evaluate seven planners
  across task difficulties and household activities. Controlled comparisons
  reveal differences in optimal planning that safe completion alone does not
  capture. Independent evaluation further identifies execution violations and
  inefficiencies in resource use and scheduling.
  \item \textbf{Companion neuro-symbolic planner.} We introduce \method{}, which
  combines language grounding, history-based state reconstruction, and
  constraint solving to jointly optimize hygiene treatment and task execution
  under user priorities.
\end{itemize}

\begin{figure*}[!t]
  \centering
  \vspace*{1.40mm}
  \includegraphics[width=\textwidth]{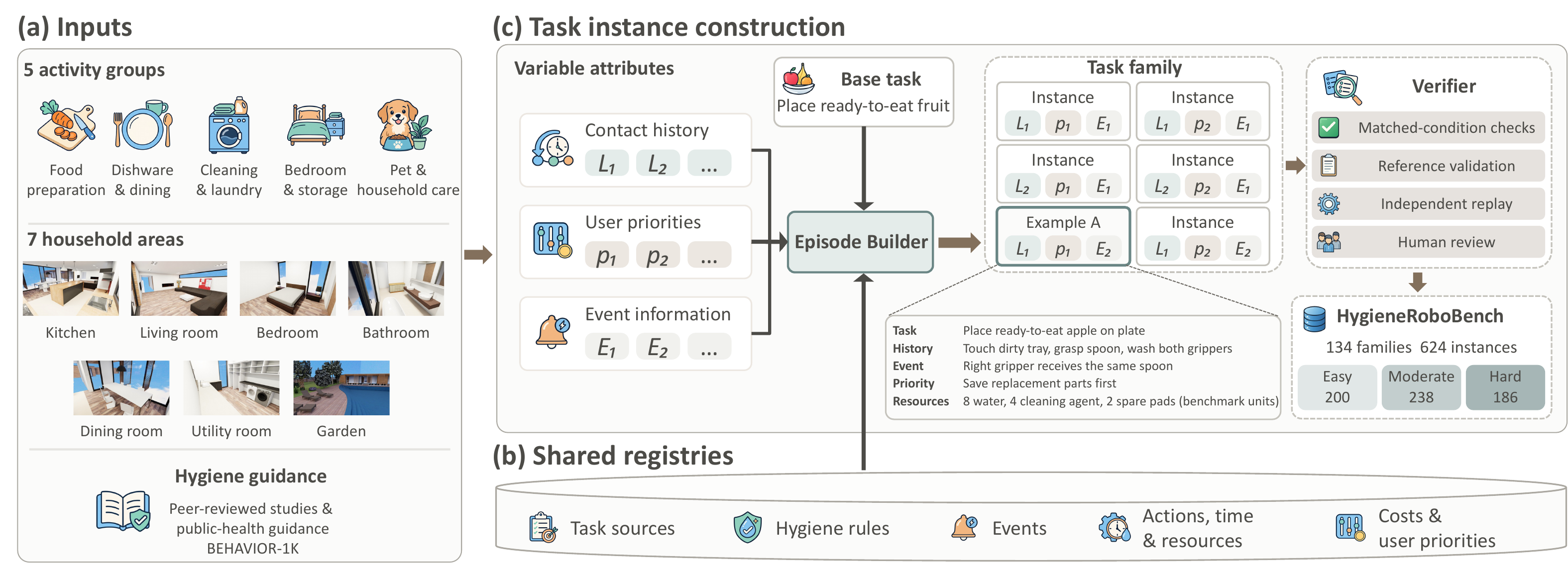}
  \caption{\textbf{Dataset construction.}
(a) Household activities and hygiene guidance supply construction inputs.
(b) Five shared registries define task, hygiene, event, resource, and user-priority specifications.
(c) Episode Builder combines a base task with contact history, user priorities,
event information, and applicable shared definitions to construct controlled
instances within a task family.
The Verifier groups matched-condition checks, Reference validation, independent replay, and human review for dataset admission.}
  \label{fig:episode_construction}
\end{figure*}

\section{Related Work}
\label{sec:related_problem}

\noindent\textbf{Task and safety benchmarks.}
BEHAVIOR-1K and RoboCasa provide household activity definitions and
manipulation environments~
\cite{li2023behavior1k,nasiriany2024robocasa}.
EmbodiedBench, Embodied Agent Interface, and PlanBench assess agents' ability
to interpret tasks, plan actions, and reason about their effects~
\cite{yang2025embodiedbench,li2024eai,valmeekam2023planbench}.
SafeAgentBench evaluates safe task planning~\cite{yin2024safeagentbench}.
IS-Bench evaluates interaction risks and their mitigation order~\cite{lu2026isbench}.
VestaBench studies long-horizon planning with multiple constraints and
adversarial settings~\cite{sadhu2025vestabench}.
SafeManip checks temporal safety properties of manipulation traces~\cite{huang2026safemanip}.
SIMMER models contamination and other latent failures in complete cooking plans~
\cite{lu2026simmer}. \benchmark{} tests how past contact changes the safe
ways to complete a task under new conditions.

\noindent\textbf{Adaptive and personalized planning.}
Inner Monologue uses environmental feedback to revise robot plans~\cite{huang2023innermonologue}.
AdaPlanBench tests plan revision as world and user constraints are revealed~\cite{liu2026adaplanbench}.
PARTNR evaluates human--robot collaboration under temporal and capability constraints~\cite{chang2025partnr}.
PbP learns user preferences from demonstrations~\cite{xu2025pbp}.
MEMENTO uses past interactions to interpret personalized requests~\cite{kwon2025memento}.
\benchmark{} tests whether planners follow given user priorities when choosing
among plans that satisfy hygiene and service constraints.

\noindent\textbf{Language and constraint-based planning.}
SayCan grounds action selection in robot affordances~\cite{ichter2023saycan}.
ProgPrompt and Code as Policies generate programmatic plans and policies~
\cite{singh2023progprompt,liang2023codepolicies}.
LLM+P connects language interpretation to formal planning~\cite{liu2023llmp}.
VIRF verifies generated plans~\cite{wu2026virf}, while ISR-LLM uses validator
feedback to refine them~\cite{zhou2024isrllm}.
SayCanPay guides cost-effective search with feasibility and payoff~\cite{hazra2024saycanpay}.
JIT-Memory maintains safety-relevant information across embodied
interaction~\cite{sima2026jitmemory}.
Finite-trace temporal logic specifies execution properties~\cite{degiacomo2013ltlf}.
Scheduling research addresses shared-resource limits and revises schedules after
disruptions~\cite{hartmann2022rcpsp,sabuncuoglu2000reactive}.
In household hygiene planning, treatment changes both action safety and available resources.
\method{} uses history-derived hygiene conditions to jointly plan treatment and
household actions under user priorities.

\section{HygieneRoboBench}
\label{sec:benchmark}

Safely completing household tasks while meeting user needs requires planners to account for contact history, user priorities, and event information. Evaluating these interrelated requirements calls for a unified benchmark that assesses plan feasibility and alignment with user priorities.

\subsection{Task Formulation}
\label{sec:task_formulation}

Each \benchmark{} \emph{task instance} requires a robot to safely fulfill a
household request within time and resource limits, following the user's
priorities among feasible plans.

Planning starts with a natural-language request $u$ and an execution history
$L_{\leq t}$ recording actions, contacts, treatments, and observed events up to decision time $t$. The robot
also receives structured physical observations $x_t$ and its remaining
resources $R_t$, which describe the current setting for completing the
request. Together with event information, user priorities, and shared rules,
these inputs form the public planning input at time $t$

\begin{equation}
  \mathcal{I}_t = (u, x_t, L_{\leq t}, R_t, E_{\leq t}, p, \mathcal{K}),
\end{equation}

where $E_{\leq t}$ references event information in $L_{\leq t}$ available to the planner by time $t$, and $p$ specifies the user's priority order over execution costs.

All planners receive the same applicable public rules and skill definitions
$\mathcal{K}$, specifying hygiene requirements and action parameters,
preconditions, effects, durations, resource use, and costs.

Given this input, the planner uses execution history to determine whether
and how the remaining task can be completed safely. The output plan $\pi_t$ covers
the remaining task. It specifies each action's objects, gripper assignment,
and start and end times in abstract time units (TU). The planner also reports
estimated execution costs. When no plan is returned, the planner reports infeasibility
or an undetermined outcome.

Feasible plans safely complete the remaining task under action, time, and
resource constraints, keeping executed actions fixed and respecting ongoing
processes. Within the feasible set $\mathcal{F}_t$, the objective is to minimize
execution costs in the lexicographic priority order specified by $p$.

\begin{figure}[tp]
  \centering
  \includegraphics[width=0.75\columnwidth]{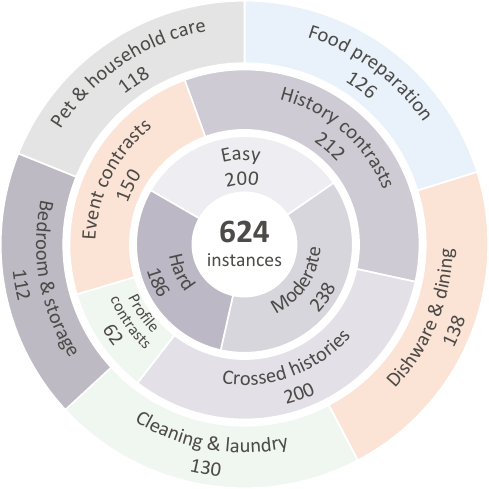}
  \caption{\textbf{Dataset composition.} The 134 task families comprise 624 instances. From outer to inner, the rings show activity groups, construction designs, and instance difficulty. Each ring independently partitions the same 624 instances, with labels indicating instance counts.}
  \label{fig:dataset_composition}
\end{figure}

\subsection{Dataset Construction}
\label{sec:construction}

Fig.~\ref{fig:episode_construction}(a)--(c) shows inputs, shared registries,
and controlled task instance construction, respectively.
Fig.~\ref{fig:dataset_composition} summarizes the instance distribution by activity, construction design, and difficulty.

\paragraph{Inputs}
We adapt BEHAVIOR-1K activities into \emph{base tasks}, preserving their
service goals, functional object roles, and necessary action ordering.
The tasks span five activity groups and seven household areas.
The hygiene-rule registry draws on peer-reviewed studies and public-health guidance~\cite{usda2019clean,fdaallergen,cdc2025home}.
We formalize these principles as contact-transfer and treatment rules with explicit material and procedure requirements.

\paragraph{Shared Registries}

Five shared registries define task requirements, action specifications, and execution costs for task construction and evaluation.

The \emph{task-source registry} anchors each adapted goal to its source activity. Given that goal, the \emph{hygiene-rule registry} defines contamination propagation, protected objects, and treatments. The \emph{action--time--resource registry} specifies action preconditions, durations, and resource requirements. The \emph{event registry} describes new contacts and other changes that update these conditions. Together, these definitions provide a common basis for determining which continuations are safe and executable.

Within this feasible set, the \emph{cost--profile registry} provides a common basis for comparing plans. It derives the execution-cost vector $\mathbf c(\pi)$ from the actions and schedule of plan $\pi$. The vector has ten dimensions: water, cleaning agent, disinfectant, replacement parts, discarded items, solid waste, energy, tool wear, active robot time, and makespan. Each user profile specifies a priority order $p$ over these cost dimensions. For the same task, changing $p$ can change how plans are ranked while leaving action costs fixed. In the fruit example, hygiene and action requirements determine whether washing or replacing the right contact pads is feasible. User priorities determine whether saving replacement parts or saving water takes precedence. Applicable entries from all five registries form the public definitions $\mathcal K$, keeping task construction, planning, and evaluation consistent.

We use an LLM to screen candidate action durations and consumable requirements from online sources.
Human reviewers check their plausibility and relative magnitudes.
We express these parameters in abstract units for consistent plan comparisons.
Examples of the shared registries are shown in the supplementary video.

\paragraph{Controlled Task Instances}
\emph{Episode Builder} combines a base task with contact history $L$, user
priorities $p$, and event information $E$.
It produces a task instance, its public input $\mathcal I_t$, and private assessment information.
For the same base task, we combine different $L$, $p$, and $E$ according to
the controlled comparison design to form a \emph{task family}.
The middle ring of Fig.~\ref{fig:dataset_composition} reports instance counts by construction type.
Crossed-history groups combine relevant and irrelevant contact variations
in a $2\times2$ design, accounting for 200 instances.
Event contrasts include no-event, relevant-event, and irrelevant-event conditions.
Of the 24 instances with a prior violation, 10 admit a safe continuation and 14 have no feasible continuation.

For difficulty analysis, we use seven instance-level structural features:
request composition, remaining-goal composition, goal-preservation conflicts,
temporal coupling, active hazards on multiple carriers, constrained treatment
resources, and relevant events known to the planner. These features define three tiers:
easy, moderate, and hard. The inner ring of Fig.~\ref{fig:dataset_composition} reports the tier distribution.

Dataset admission combines matched-condition checks, Reference validation of
feasibility and costs, independent full-history replay, and human review.
Three reviewers assess the entire dataset for task wording,
household plausibility, and treatment applicability.

\subsection{Evaluation Protocol and Metrics}
\label{sec:evaluation_protocol}
We evaluate submitted continuation plans and judgments of infeasibility.
The Evaluator checks submitted plans through full-history replay, independently of the planning solver.
It checks whether the submitted plan achieves the goals and satisfies hygiene, timing, and resource constraints.
It independently computes execution costs from the plan's actions and schedule.
The Reference provides feasibility and certified optimal
costs under the user's priorities within the declared action and time bounds.

Planners receive only the public input $\mathcal I_t$. Every response,
including one without a plan, is assessed against private Reference targets.
These targets and evaluation feedback remain hidden from planners.

For each contact-event instance, all planners receive the same updated history and task context.
We report SR and OSR separately for relevant and irrelevant events, using the same instances for all planners.
These instances have safe prefixes, certified-feasible continuations, and certified reference optima.

\paragraph{Safe Resolution}
Safe Resolution Rate (SR) measures safe completion or correct rejection of
the remaining task from the prescribed decision point:
\begin{equation}
  \mathrm{SR}=\tfrac{1}{N}\sum_{i\in\mathcal{D}}
  \bigl[f_i b_i+(1-f_i)r_i\bigr].
\end{equation}
Here, $\mathcal{D}$ is the common comparison set and $N=|\mathcal{D}|$.
The dataset contains 604 instances with certified-feasible continuations and
20 with certified-infeasible continuations. The indicator $f_i$ denotes
continuation feasibility, and $r_i$
denotes an explicit infeasibility judgment without a submitted plan.
Safe completion is $b_i=v_i$, where $v_i$ denotes a submitted continuation satisfying the goal, hygiene, timing,
resource, and other feasibility conditions in Section~\ref{sec:task_formulation}.
Earlier violations remain recorded but do not by themselves determine continuation feasibility.
Undetermined outcomes and incorrect rejections do not count as resolutions.

\paragraph{User Priorities and Execution Costs}
Under user priorities $p$, we compare recomputed plan costs with the
certified reference optimum. Optimal Safe Resolution Rate (OSR) requires safe
completion at this optimum or correct rejection of a certified-infeasible task.
Profile pairs vary only user priorities ($p_1$ versus $p_2$), with history $L$,
event information $E$, and other task conditions fixed.
Profile-Pair Success (PS) counts pairs where both instances complete safely
at their respective reference-optimal costs.
\begin{equation}
\begin{aligned}
  \mathrm{OSR}&=\frac{1}{N}\sum_{i\in\mathcal{D}}
      \bigl[f_i b_i o_i m_i+(1-f_i)r_i\bigr],\\
  \mathrm{PS}&=
      \sum_{(i,j)\in\mathcal{U}}b_i b_j m_i m_j.
\end{aligned}
\end{equation}
Here, $o_i$ indicates a certified reference optimum under instance $i$'s
priorities $p$. For submitted plan $\pi_i$, $m_i=1$ when its recomputed cost
$\mathbf c(\pi_i)$ matches that optimum, and zero when no match is established.
The set $\mathcal U$ contains profile pairs whose members both have safe
prefixes and certified-feasible continuations.
We report PS on general profile pairs and on ranking-flip and ranking-invariant controls.
These controls are defined by whether changing user priorities changes the cost ranking of verified alternative plans.

\paragraph{History Response}
History pairs vary designated contact records in $L$ while keeping
event information $E$, user priorities $p$, and other task conditions fixed.
History-pair completion (HC) counts the pairs in which both instances complete safely.
HC$^{*}$ additionally requires both to match their respective
reference-optimal costs:
\begin{equation}
  \mathrm{HC}=
      \sum_{(i,j)\in\mathcal{H}}b_i b_j,\quad
  \mathrm{HC}^{*}=
      \sum_{(i,j)\in\mathcal{H}}b_i b_j m_i m_j.
\end{equation}
Both metrics use the same pair set $\mathcal H$, with each member required
to have a safe prefix and a certified-feasible continuation.
In this benchmark, relevant-history pairs admit a shared safe continuation
but have different certified optimal cost vectors. They therefore require
different optimal continuations. Irrelevant-history pairs admit a shared
certified-optimal continuation under the same task conditions.

\section{Hygiene-NSP: A Hygiene-Aware Neuro-Symbolic Planner}
\label{sec:method}

Hygiene-NSP reconstructs hygiene conditions from contact history and jointly
plans treatment and service actions under user priorities. These decisions
are coupled: a washed gripper can be contaminated again, and cleaning or
replacement consumes resources needed for service. Algorithm~\ref{alg:hygiene_nsp}
summarizes one planning call, repeated on the updated public input after
an action or an update to event information.

\begingroup
\makeatletter
\let\hygiene@ruled\fs@ruled
\def\fs@ruled{\hygiene@ruled
  \let\hygiene@floatpre\@fs@pre
  \def\@fs@pre{\vskip1.40mm\hygiene@floatpre}}
\makeatother
\begin{algorithm}[t]
\caption{Hygiene-NSP continuation planning}
\label{alg:hygiene_nsp}
\footnotesize
\begin{algorithmic}[1]
\Require Public input $\mathcal I_t$, solver time budget $B$
\Ensure Plan and modeled costs, or infeasible/undetermined
\State $\mathcal P_t \gets \operatorname{Compile}_{\mathcal K}(\operatorname{Ground}(\mathcal I_t))$
\State $S_t \gets \operatorname{Replay}_{\mathcal K}(L_{\leq t})$; $\bar S_t \gets S_t$
\State \textbf{if} projection is enabled and supported \textbf{then}
\State \hspace{\algorithmicindent}$S'_t \gets \operatorname{Project}(S_t,\mathcal P_t)$
\State \hspace{\algorithmicindent}\textbf{if} $\operatorname{Consistent}(S'_t,S_t)$ \textbf{then} $\bar S_t \gets S'_t$
\State $\widehat{\mathcal F}_t \gets \operatorname{JointModel}(\mathcal P_t,\bar S_t)$
\State $(q,\pi_t,\mathbf c) \gets \operatorname{LexCP\text{-}SAT}(\widehat{\mathcal F}_t,p,B)$
\State \Return $(q,\pi_t,\mathbf c)$
\end{algorithmic}
\smallskip
{\footnotesize $q$ distinguishes success, proved infeasibility, and undetermined outcomes.}
\end{algorithm}
\endgroup

\subsection{Language Grounding and Constraint Construction}

The LLM Grounder reads the request and lists the goals and their types, identifies the objects involved, and records the time requirements. It also includes the supplied user priorities and event information. The recorded observations and resource amounts keep their given values.

The Trusted Compiler converts these goals and the shared rules $\mathcal K$ into a planning problem $\mathcal P_t$. It checks that object names match available objects, sets the requirements and effects of each action, and updates the affected constraints using event information.

\subsection{History Replay and State Reconstruction}

Physical observations alone need not indicate the hygiene consequences of earlier actions. Moreover, a record that an object was washed is insufficient if later contact can contaminate it again. The planner derives the current hygiene state from the order and effects of past contacts and treatments.

We start with the given initial conditions and follow the recorded actions in $L_{\leq t}$ in order. The shared rules $\mathcal K$ specify how contamination from each source spreads and how completed treatments change it. This process, called history replay, gives the hygiene state for the remaining task:

\begin{equation}
  S_t=\operatorname{Replay}_{\mathcal K}(L_{\leq t}).
  \label{eq:history_recovery}
\end{equation}

History replay reconstructs the current hygiene state while preserving past violations and cumulative resource consumption. After another action or a new event, the planner updates the planning problem and replays the full history before planning the continuation.

Optional state projection retains state information and past evidence relevant to the remaining task. A checker validates the retained states, past violations, and cumulative resource consumption against full replay before the smaller state $\bar S_t$ is used. With projection disabled or its support conditions unestablished, planning uses the full state, $\bar S_t=S_t$.

\begin{table*}[!t]
  \vspace*{1.65mm}
  \caption{\textbf{Overall performance and breakdowns by instance difficulty and activity.}}
  \label{tab:main}
  \centering\footnotesize
  \setlength{\tabcolsep}{2pt}\renewcommand{\arraystretch}{1.06}
  \textbf{A. Overall performance and instance difficulty (\%)}\par\vspace{3pt}
\begin{tabular*}{\textwidth}{@{\extracolsep{\fill}}lrrrrrrrr@{}}
\toprule
& \multicolumn{2}{c}{Overall ($N=624$)} & \multicolumn{2}{c}{Easy ($n=200$)} & \multicolumn{2}{c}{Moderate ($n=238$)} & \multicolumn{2}{c}{Hard ($n=186$)} \\
\cmidrule(lr){2-3}\cmidrule(lr){4-5}\cmidrule(lr){6-7}\cmidrule(l){8-9}
Planner & SR & OSR & SR & OSR & SR & OSR & SR & OSR \\
\midrule
GPT-5.6-Sol & 78.5 & 76.9 & 87.5 & 85.5 & 78.6 & 77.7 & 68.8 & 66.7 \\
GPT-5.6-Luna & 46.6 & 37.0 & 55.0 & 45.5 & 52.1 & 41.2 & 30.6 & 22.6 \\
Gemini 3.6 Flash & 89.7 & \underline{86.1} & 95.0 & \underline{91.5} & 90.8 & \underline{88.7} & \underline{82.8} & \underline{76.9} \\
DeepSeek-V4.1-Flash & \underline{90.7} & 75.0 & \underline{96.0} & 83.5 & \underline{92.4} & 71.8 & \underline{82.8} & 69.9 \\
MiniMax-M3 & 67.0 & 57.2 & 74.5 & 66.0 & 70.2 & 58.8 & 54.8 & 45.7 \\
Rule-SAT & 30.9 & 30.0 & 38.0 & 37.0 & 36.6 & 35.3 & 16.1 & 15.6 \\
Hygiene-NSP & \textbf{94.4} & \textbf{90.4} & \textbf{96.5} & \textbf{96.0} & \textbf{96.2} & \textbf{91.6} & \textbf{89.8} & \textbf{82.8} \\
\bottomrule\end{tabular*}
\par\vspace{7pt}
\textbf{B. Performance by activity (\%)}\par\vspace{3pt}
\begin{tabular*}{\textwidth}{@{\extracolsep{\fill}}lrrrrrrrrrr@{}}
\toprule
& \multicolumn{2}{c}{\shortstack{Food\\preparation\\($n=126$)}} & \multicolumn{2}{c}{\shortstack{Dishware\\\& dining\\($n=138$)}} & \multicolumn{2}{c}{\shortstack{Cleaning\\\& laundry\\($n=130$)}} & \multicolumn{2}{c}{\shortstack{Bedroom\\\& storage\\($n=112$)}} & \multicolumn{2}{c}{\shortstack{Pet \&\\household care\\($n=118$)}} \\
\cmidrule(lr){2-3}\cmidrule(lr){4-5}\cmidrule(lr){6-7}\cmidrule(lr){8-9}\cmidrule(l){10-11}
Planner & SR & OSR & SR & OSR & SR & OSR & SR & OSR & SR & OSR \\
\midrule
GPT-5.6-Sol & 76.2 & 73.8 & 73.9 & 69.6 & 84.6 & 83.8 & 75.9 & 75.9 & 82.2 & 82.2 \\
GPT-5.6-Luna & 38.1 & 29.4 & 44.9 & 33.3 & 54.6 & 46.9 & 34.8 & 31.2 & 60.2 & 44.1 \\
Gemini 3.6 Flash & \underline{86.5} & \underline{85.7} & 87.7 & \underline{84.1} & \textbf{92.3} & \textbf{86.2} & \underline{91.1} & \textbf{88.4} & 91.5 & \underline{86.4} \\
DeepSeek-V4.1-Flash & 83.3 & 65.9 & \underline{88.4} & 64.5 & \underline{91.5} & 79.2 & \textbf{96.4} & \underline{86.6} & \underline{94.9} & 81.4 \\
MiniMax-M3 & 69.0 & 60.3 & 58.7 & 50.7 & 75.4 & 61.5 & 67.9 & 63.4 & 64.4 & 50.8 \\
Rule-SAT & 38.1 & 34.9 & 15.9 & 15.9 & 37.7 & 36.9 & 37.5 & 36.6 & 27.1 & 27.1 \\
Hygiene-NSP & \textbf{100.0} & \textbf{100.0} & \textbf{89.9} & \textbf{86.2} & 86.9 & \underline{85.4} & \textbf{96.4} & 82.1 & \textbf{100.0} & \textbf{98.3} \\
\bottomrule\end{tabular*}

  \par\vspace{3pt}
  \begin{minipage}{\textwidth}\scriptsize
Activity headers give instance counts. Across Tables~\ref{tab:main}, \ref{tab:paired_results}, and~\ref{tab:event_results}, bold and underlining indicate the best and second-best distinct values for each metric.
  \end{minipage}
\end{table*}

\subsection{Joint Hygiene and Service Optimization}

A necessary hygiene treatment is also an action that consumes time and resources. Fixing a treatment before scheduling the service can exclude alternatives that better satisfy the remaining constraints or user priorities. We therefore choose treatment routes, service actions, and their timing jointly, first enforcing feasibility and then comparing feasible plans by their execution costs.

CP-SAT~\cite{perron2023cpsatlp} uses $\mathcal P_t$ and $\bar S_t$ to choose actions and their start and end times. Only selected actions activate their time intervals and update task and hygiene states according to the rules. Other rules prevent conflicting uses of the same gripper, keep simultaneous resource use within shared capacity, and keep total consumption within the remaining supplies. Starting with $\bar S_t$, the model tracks contamination to prevent it from reaching protected objects. These constraints define $\widehat{\mathcal F}_t$, the feasible set of Hygiene-NSP's grounded planning model. The benchmark feasible set $\mathcal F_t$ follows the task requirements in Section~\ref{sec:task_formulation}. The independent Evaluator checks submitted plans against those requirements.

The solver optimizes costs in priority order $p$, fixing each proved optimum before proceeding to the next component:

\begin{equation}
\begin{aligned}
  \gamma_j &= \min_{\pi\in\widehat{\mathcal F}_t} c_{p_j}(\pi),\\
  \text{subject to}\quad
  c_{p_\ell}(\pi)&=\gamma_\ell \quad \text{for all } \ell<j.
\end{aligned}
  \label{eq:staged_optimization}
\end{equation}
Here, $p_j$ selects the next component of $\mathbf c(\pi)$, and $\gamma_\ell$
is a previously proved stage optimum.
If optimization remains incomplete, the solver retains any feasible plan found
without claiming full lexicographic optimality. If no plan is found and
infeasibility is not proved, it returns an undetermined outcome.
When optimization succeeds, it returns $\pi_t$ and its modeled cost vector for independent evaluation.

\section{Experiments}
\label{sec:experiments}

We evaluate overall resolution (RQ1), plan choices under different histories
and user priorities (RQ2), and safe, cost-optimal continuation after contact events
(RQ3). A separate method analysis examines state projection.

\subsection{Experimental Setup}

We evaluated GPT-5.6-Sol, GPT-5.6-Luna, Gemini 3.6 Flash,
DeepSeek-V4.1-Flash, and MiniMax-M3 on 624 instances across 134 task families,
alongside Rule-SAT and \method{}. The language models generate complete plans
directly. Rule-SAT constructs constraint problems through rule-based parsing,
while \method{} uses GPT-5.6-Sol as its language grounder. Both planners use CP-SAT.
We used high reasoning settings where available and enabled thinking for MiniMax-M3.

All planners receive the same public input $\mathcal I_t$ and applicable
rules $\mathcal K$, and their outputs are evaluated under the protocol in
Section~\ref{sec:evaluation_protocol}. The main comparison uses full contact
history. Event response is assessed on the event subset of this dataset,
and state projection is examined separately.

\subsection{RQ1: How Reliably Do Planners Resolve Requests?}
\label{sec:rq1}

We evaluate safe resolution (SR) and optimal safe resolution (OSR) under user priorities.
Both metrics credit correct rejection of infeasible continuations.
Table~\ref{tab:main} reports
overall results, difficulty tiers, and activity breakdowns.

\textbf{\benchmark{} distinguishes planner performance across task difficulty and household activities.}
\method{} achieves the highest overall SR and OSR, at 94.4\% and 90.4\%,
respectively (Table~\ref{tab:main}-A). SR and OSR decrease from easy to moderate
to hard instances for all seven planners. This consistent trend shows that
\benchmark{}'s difficulty tiers, defined by task structure, distinguish planner
performance at different levels of task complexity. Activity-level results
expose differences hidden by the overall ranking: Gemini 3.6 Flash achieves
the highest SR and OSR in cleaning and laundry (Table~\ref{tab:main}-B).

\begin{table}[!htb]
  \caption{\textbf{Failure and cost outcomes across 624 instances per planner.}
  Counts are mutually exclusive. The first three columns fail SR.
  The last passes SR but not OSR.}
  \label{tab:failure_cost}
  \centering\footnotesize
  \setlength{\tabcolsep}{1pt}\renewcommand{\arraystretch}{1.08}
  \begin{tabular*}{\linewidth}{@{\extracolsep{\fill}}lrrrr@{}}
\toprule
& \shortstack{Other\\no plan} & \shortstack{Incorrect\\rejection} & \shortstack{Invalid\\continuation} & \shortstack{Safe,\\nonoptimal} \\
\midrule
GPT-5.6-Sol & 1 & 25 & 108 & 10 \\
GPT-5.6-Luna & 27 & 112 & 194 & 60 \\
Gemini 3.6 Flash & 8 & 12 & 44 & 23 \\
DeepSeek-V4.1-Flash & 3 & 7 & 48 & 98 \\
MiniMax-M3 & 33 & 10 & 163 & 61 \\
Rule-SAT & 415 & 12 & 4 & 6 \\
Hygiene-NSP & 22 & 13 & 0 & 25 \\
\bottomrule
\end{tabular*}

\end{table}

\textbf{\benchmark{} identifies execution and hygiene violations that checks of service-goal completion alone can miss.}
\method{} produces no invalid continuations in the full evaluation
(Table~\ref{tab:failure_cost}). In a leftover-storage task, DeepSeek-V4.1-Flash
completes the requested placements. However, it grasps a protected food container
with a gripper exposed to peanut residue through a shared handle.
Full-history replay detects the resulting contamination. On the same instance,
\method{} replaces the contact insert before moving and placing the objects.
Its plan is safe and matches the reference-optimal cost.

\textbf{Independent cost evaluation identifies resource and scheduling inefficiencies in safe plans.}
DeepSeek-V4.1-Flash returns 98 safe, nonoptimal plans (Table~\ref{tab:failure_cost}).
In 41 of these plans, the first nonoptimal cost dimension under user priorities
is water or replacement parts. In a hot-cocoa task, DeepSeek-V4.1-Flash and
\method{} use the same actions and consumables. DeepSeek-V4.1-Flash places the
marshmallows after cooking and pouring the cocoa, taking 19 TU in total.
\method{} places them during cooking and finishes in 13 TU, matching the
reference optimum.

\subsection{RQ2: Do Plans Adapt to History and User Priorities?}
\label{sec:rq2}

Controlled pairs test whether planners select safe, optimal continuations
under different histories and user priorities.

The 62 profile-contrast instances form 31 pairs. Two pairs have prior violations
and do not meet the safe-prefix requirement, leaving 24 ranking-flip and five
ranking-invariant pairs. The spoon--fruit task family contributes one additional
ranking-flip pair.

\begin{table}[!htb]
  \caption{\textbf{Controlled history and priority pairs.} Entries give HC/HC$^{*}$ or PS counts; $n$ counts eligible pairs from the construction groups in Fig.~\ref{fig:dataset_composition}. Evaluation sets may share instances.}
  \label{tab:paired_results}
  \centering\footnotesize
  \setlength{\tabcolsep}{1.5pt}\renewcommand{\arraystretch}{1.12}
  \textbf{A. History pairs (HC / HC$^{*}$)}\par
\begin{tabular*}{\linewidth}{@{\extracolsep{\fill}}lcc@{}}
\toprule
Planner & Relevant ($n=180$) & Irrelevant ($n=92$) \\
\midrule
GPT-5.6-Sol & 126 / 123 & 69 / 66 \\
GPT-5.6-Luna & 57 / 38 & 27 / 16 \\
Gemini 3.6 Flash & 157 / \underline{141} & \underline{81} / \underline{76} \\
DeepSeek-V4.1-Flash & \underline{162} / 118 & \textbf{86} / 60 \\
MiniMax-M3 & 94 / 65 & 51 / 33 \\
Rule-SAT & 74 / 71 & 50 / 50 \\
\midrule
Hygiene-NSP & \textbf{168} / \textbf{156} & \textbf{86} / \textbf{78} \\
\bottomrule\end{tabular*}
\par\smallskip\textbf{B. Profile pairs (PS)}\par
\begin{tabular*}{\linewidth}{@{\extracolsep{\fill}}lccc@{}}
\toprule
Planner & \shortstack{Ranking flip\\($n=25$)} & \shortstack{Ranking invariant\\($n=5$)} & \shortstack{General\\($n=268$)} \\
\midrule
GPT-5.6-Sol & 20 & \underline{3} & 184 \\
GPT-5.6-Luna & 7 & \underline{3} & 58 \\
Gemini 3.6 Flash & \underline{23} & \textbf{5} & \underline{218} \\
DeepSeek-V4.1-Flash & 16 & \underline{3} & 178 \\
MiniMax-M3 & 11 & \underline{3} & 108 \\
Rule-SAT & 6 & 0 & 83 \\
\midrule
Hygiene-NSP & \textbf{24} & \underline{3} & \textbf{234} \\
\bottomrule\end{tabular*}

\end{table}

\textbf{History-pair evaluation identifies differences in optimal planning that safe completion counts alone do not capture.}
On relevant-history pairs, \method{} achieves the highest HC$^{*}$,
at 156 (Table~\ref{tab:paired_results}-A). DeepSeek-V4.1-Flash has a higher HC
than Gemini 3.6 Flash but a lower HC$^{*}$. Including optimality therefore
reverses their ranking. On irrelevant-history pairs, \method{} and
DeepSeek-V4.1-Flash both achieve an HC of 86, but their HC$^{*}$ counts are
78 and 60, respectively.

We further evaluated saved planner outputs under history swaps across the
180 relevant-history and 92 irrelevant-history pairs. We tested both
directions and kept submitted actions and schedules unchanged.
Across seven planners, none of the
1,699 originally safe, optimal plan replays remained both safe and optimal
under relevant-history swaps. All 1,288 irrelevant-history swap evaluations
preserved the original safety and optimality outcomes. These results support
the validity of the controlled history pairs: relevant-history changes affect
plan safety or optimality, whereas irrelevant-history changes preserve the
evaluated outcomes.
The supplementary video presents the swap evaluation procedure and results by planner.

\textbf{User-priority pairs evaluate safe, optimal planning under different resource preferences.}
\method{} succeeds on 234 of 268 general profile pairs and 24 of 25
ranking-flip pairs, the highest counts in both groups
(Table~\ref{tab:paired_results}-B). In the spoon--fruit task illustrated in
Fig.~\ref{fig:motivation}, \method{} switches between washing and replacement
under the two priority profiles, attaining the reference optimum in both cases.

\subsection{RQ3: Can Planners Continue Safely at Optimal Costs After Contact Events?}
\label{sec:rq3}

We evaluate 50 relevant-event and 50 irrelevant-event instances across
25 task families. Table~\ref{tab:event_results} reports SR and OSR for both conditions.

\begin{table}[!htb]
  \caption{\textbf{Contact-event results.} Entries report SR/OSR (\%); each condition contains 50 instances.}
  \label{tab:event_results}
  \centering\footnotesize
  \setlength{\tabcolsep}{3pt}
  \begin{tabular}{@{}lcc@{}}
\toprule
Planner & Relevant SR/OSR & Irrelevant SR/OSR \\
\midrule
GPT-5.6-Sol & 78/74 & 74/70 \\
GPT-5.6-Luna & 40/24 & 34/30 \\
Gemini 3.6 Flash & \textbf{92}/\textbf{90} & \textbf{94}/\textbf{94} \\
DeepSeek-V4.1-Flash & 88/66 & \underline{92}/86 \\
MiniMax-M3 & 76/62 & 62/56 \\
Rule-SAT & 12/12 & 12/12 \\
\midrule
Hygiene-NSP & \underline{90}/\underline{88} & \underline{92}/\underline{92} \\
\bottomrule
\end{tabular}

\end{table}

\textbf{\method{} achieves high safe and optimal safe resolution rates under both contact-event conditions.}
Under relevant and irrelevant events, its SR is 90\% and 92\%, respectively,
while its OSR is 88\% and 92\%. DeepSeek-V4.1-Flash shows a larger difference
between the two conditions, with OSR of 66\% and 86\%.
Gemini 3.6 Flash achieves the highest OSR in both conditions.

\method{} and Gemini exhibit different failure types. Across the 100
contact-event instances, \method{}'s nine SR failures comprise five
undetermined outcomes and four incorrect rejections. All of its submitted
plans pass execution checks. Gemini's seven SR failures comprise four
incorrect rejections and three invalid continuations.

\subsection{Method Analysis: State Projection}
\label{sec:projection_analysis}

We compared full-state (Full) and projected-state (Projected) planning on
59 task families with three paired runs each. The median family-level
reduction in local planning time was 20.1\%, excluding LLM calls.
Across 177 pairs, including 21 full-state fallbacks, plan status and cost
agreed in 175 pairs, with no new hygiene violations. In one differing pair,
Projected found a safe plan where Full returned undetermined. In the other,
it reduced makespan from 17 to 10 TU relative to Full's feasible plan.

\section{Conclusion and Future Work}
\label{sec:conclusion}

We introduced \benchmark{}, with 624 instances across 134 task families.
Controlled construction and independent evaluation assess household service
planning under contact history, user priorities, and contact events.
All seven planners perform worse as structural task difficulty increases,
demonstrating the benchmark's ability to distinguish performance across task
complexity. Controlled pairs further reveal differences in optimal planning
that safe completion counts alone do not capture. \method{} achieves the
highest overall SR and OSR, at 94.4\% and 90.4\%, respectively, with no invalid
submitted continuations.

\benchmark{} provides a common basis for evaluating hygiene decisions within
complete service tasks, connecting contact-history reasoning with resource
trade-offs under user priorities. The current evaluation uses explicit contact
and treatment rules. Future work will extend evaluation to physical execution,
using measured contact and treatment outcomes to assess these planning decisions.

\bibliographystyle{IEEEtran}
\bibliography{refs}
\end{document}